\def\year{2022}\relax
\documentclass[letterpaper]{article} 
\usepackage{aaai22}  
\usepackage{times}  
\usepackage{helvet}  
\usepackage{courier}  
\usepackage[hyphens]{url}  
\usepackage{graphicx} 
\usepackage{natbib}  
\usepackage{caption} 
\usepackage{xcolor,colortbl}

\usepackage{color}
\DeclareCaptionStyle{ruled}{labelfont=normalfont,labelsep=colon,strut=off} 
\usepackage{algorithm}
\usepackage{algorithmic}
\usepackage{amssymb}
\usepackage{amsmath}
\usepackage{bbding}
\usepackage{newfloat}
\usepackage{listings}
\floatstyle{ruled}
\newfloat{listing}{tb}{lst}{}
\floatname{listing}{Listing}
\usepackage{url}
\usepackage[pagebackref=true,breaklinks=true,letterpaper=true,colorlinks,bookmarks=false]{hyperref}
\title{Coarse-to-Fine Embedded PatchMatch and Multi-Scale Dynamic Aggregation for Reference-based Super-Resolution }
\author {
	Bin Xia\textsuperscript{\rm 1},
	Yapeng Tian\textsuperscript{\rm 2}, 
	Yucheng Hang\textsuperscript{\rm 1}, 
	Wenming Yang\textsuperscript{\rm 1}\thanks{Corresponding  author.}, Qingmin Liao\textsuperscript{\rm 1}, Jie Zhou\textsuperscript{\rm 1}
}
\affiliations{
	\textsuperscript{\rm 1} Tsinghua University\\
	\textsuperscript{\rm 2} University of Rochester\\
	
	xiab20@mails.tsinghua.edu.cn, yapengtian@rochester.edu, hangyc20@mails.tsinghua.edu.cn,  yang.wenming@sz.tsinghua.edu.cn, liaoqm@tsinghua.edu.cn,  jzhou@tsinghua.edu.cn
}

\usepackage{bibentry}

\begin{document}

\maketitle

\begin{abstract}
Reference-based super-resolution (RefSR) has made significant progress in producing realistic textures using an external reference (Ref) image. However, existing RefSR methods obtain high-quality correspondence matchings consuming quadratic computation resources with respect to the input size, limiting its application. Moreover, these approaches usually suffer from scale misalignments between the low-resolution (LR) image and Ref image. In this paper, we propose an Accelerated Multi-Scale Aggregation network (AMSA) for Reference-based Super-Resolution, including Coarse-to-Fine Embedded PatchMatch (CFE-PatchMatch) and Multi-Scale Dynamic Aggregation (MSDA) module. To improve matching efficiency, we design a novel Embedded PatchMacth scheme with random samples propagation, which involves end-to-end training with asymptotic linear computational cost to the input size. To further reduce computational cost and speed up convergence, we apply the coarse-to-fine strategy on Embedded PatchMacth constituting CFE-PatchMatch. To fully leverage reference information across multiple scales and enhance robustness to scale misalignment, we develop the MSDA module consisting of Dynamic Aggregation and Multi-Scale Aggregation. The Dynamic Aggregation corrects minor scale misalignment by dynamically aggregating features, and the Multi-Scale Aggregation brings robustness to large scale misalignment by fusing multi-scale information. Experimental results show that the proposed AMSA achieves superior performance over state-of-the-art approaches on both quantitative and qualitative evaluations. The code is available at \url{https://github.com/Zj-BinXia/AMSA}.
	
\end{abstract}

\section{Introduction}
\begin{figure}[htb]
	\centering
	\includegraphics[height=4.7cm,width=8cm]{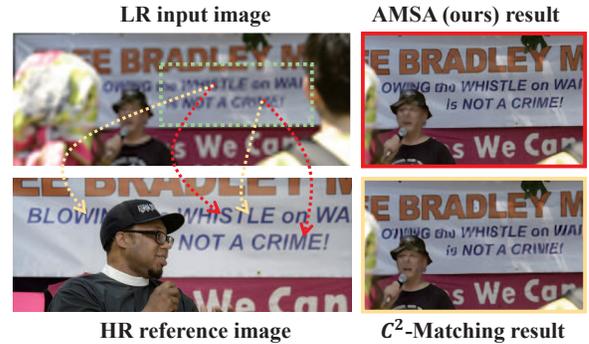} 
	\caption{Visual comparison of 4$\times$ SR results. The correspondences matched by our AMSA are marked in red, and the correspondences matched by the leading RefSR method, $C^{2}$-Matching \cite{C2Matching}, are marked in yellow. Compared with $C^{2}$-Matching, our approach can effectively match and leverage reference information from the HR reference image to reconstruct a visually appealing HR image with more accurate structures and fewer artifacts.}
	\label{fig:head}
\end{figure}
Reference-based Super-resolution (RefSR) aims to reconstruct a photo-realistic high-resolution (HR) image from the low-resolution (LR) counterpart with the
guidance of an additional HR reference (Ref) image. The reference image contains rich and diverse structures and textures, which could be relevant to the content in the target HR image. By transferring relevant information between the LR image and HR reference, recent RefSR methods \cite{{TTSR},{SRNTT},{MASA},{C2Matching}} have shown promising results.

The essential step in texture transfer for RefSR is to find correspondences between the input image and the reference image. However, existing RefSR methods \cite{{TTSR},{SRNTT}} enumerate all possible patch pairs and select the best-matched patches for aggregation, the computational complexity of which is quadratic to image size. Although MASA \cite{MASA} matches correspondences in coarser resolution, yet consumes computation resources quadratic to spatial size. Besides, since the convolution operation is sensitive to the input scale \cite{visualizing}, RefSR may miss the best patch for reconstruction while matching and aggregating relevant feature patches at different scales. To alleviate the resolution gap, $C^{2}$-Matching \cite{C2Matching} recently applied knowledge distillation to force correspondences relevant in feature space. Nevertheless, with the increase of scale misalignment, the performance of $C^{2}$-Matching decreases rapidly. 
 
To address the challenges mentioned above, we develop an Accelerated Multi-Scale Aggregation network (AMSA) for Reference-based Super-Resolution. The design of AMSA has several advantages. First, motivated by the discovery that a large number of random samples often lead to good guesses \cite{{patchmatch},{jumpflood},{deeppruner},{patchmatchnet}}, we design a novel Embedded PatchMatch scheme embedded in the network for end-to-end training, which merely requires the  asymptotic linear computational cost to the input size. To further reduce computational cost and accelerate the convergence of Embedded PatchMatch, we adopt the coarse-to-fine strategy that uses the matchings from the coarse level to provide the subsequent finer levels a good initialization start, constituting our Coarse-to-Fine Embedded PatchMatch (CFE-PatchMatch). 

Second, to enhance robustness to scale misalignment between the LR and Ref image and makes full use of multi-scale relevant information in the Ref image, we propose the Multi-Scale Dynamic Aggregation (MSDA) consisting of Dynamic Aggregation and Multi-Scale Aggregation. The Dynamic Aggregation uses the DCN \cite{Deform} form to correct minor scale misalignment. Furthermore, the Multi-Scale Aggregation fuses multi-scale reference information to obtain robustness to large scale misalignment. In sum, as shown in Figure \ref{fig:head}, compared with the state-of-the-art method \cite{C2Matching}, our AMSA matches and transfers reference information effectively to produce visually pleasant details. Additionally, we achieve over 0.3dB improvement on the CUFED5 dataset with 100 times acceleration in matching correspondences. Our main contributions are as follows:

\begin{itemize}
	\item We design a novel Coarse-to-Fine Embedded PatchMatch scheme, which involves end-to-end training successfully and obtains good matching results with asymptotic linear computational cost to the input size. 
	\item To enhance robustness to scale misalignment and fully utilize the multi-scale relevant information in the Ref image, we propose the Multi-Scale Dynamic Aggregation module. MSDA uses Dynamic Aggregation to correct minor scale misalignment and applies Multi-Scale Aggregation to obtain robustness to large scale misalignment. 
	\item Compared with SOTA methods, our model achieves superior performance with less computational cost in correspondence matching.	
\end{itemize} 

\section{Related Work}

\subsection{Single Image Super-Resolution}
The SISR methods \cite{{SRCNN},{EDSR},{VDSR},{RCAN},{SRGAN},{RDN},{secondorder}} based on deep learning learn an end-to-end image mapping function between LR and HR images, which achieves better performance than conventional algorithms. Generally, commonly used Mean Square Error (MSE) and Mean Absolute Error (MAE) objective functions are prone to generate unexpected over-smoothing results. Consequently, various subsequent works tend to design loss functions to enhance visual quality. \citet{p_loss} introduced perceptual loss with VGG \cite{vgg} and showed visually satisfying results. SRGAN \cite{SRGAN} enforced SR results close to the distribution of the natural image by taking GANs as the adversarial loss. Furthermore, the knowledge distillation framework is also explored to improve the SR performance in recent works \cite{{Knowledge_dis1},{Knowledge_dis2}}.

\subsection{Reference-based SR}
Unlike Single Image Super-Resolution (SISR), RefSR reconstructs finer details by referring to additional HR images. The Ref image generally has relevant content with the LR image to provide realistic details. Recent works \cite{{E2Net},{CIMR-SR}} mostly use CNN-based frameworks. One part of RefSR implicitly aligns Ref and LR images by the network. CrossNet \cite{CrossNet} warped reference features with optical flow. Nevertheless, it relied on the  flow estimated network, resulting in high computational burdens and inaccurate estimation. Besides, SSEN \cite{SSEN} employed deformable convolution \cite{Deform} to align the LR and Ref images. 

Another part of RefSR explicitly matches relevant features. SRNTT \cite{SRNTT} and TTSR \cite{TTSR} applied an enumerated patch matching mechanism upon the feature map to swap relevant texture features, requiring extensive computational resources. To alleviate the problem, MASA \cite{MASA} adopts a hierarchical enumerated correspondence matching scheme. Furthermore, the resolution gap between LR and Ref images has negative impacts on transferring HR details. Consequently, $C^{2}$-Matching introduced knowledge \cite{hinton2015distilling} to force correspondences of LR and Ref images to be similar.
However, the computational complexity of MASA is quadratic to the input size, and the performance of $C^{2}$-Matching still decreases sharply with the scale misalignment between LR and Ref images increasing. To address the above issues, we propose CFE-PatchMatch to further reduce computational cost and develop the MSDA module to enhance robustness to the scale misalignment and fully use informative features across multiple scales.

\section{Method}
\begin{figure*}[htb]
	\centering
	\includegraphics[height=6.8cm,width=16cm]{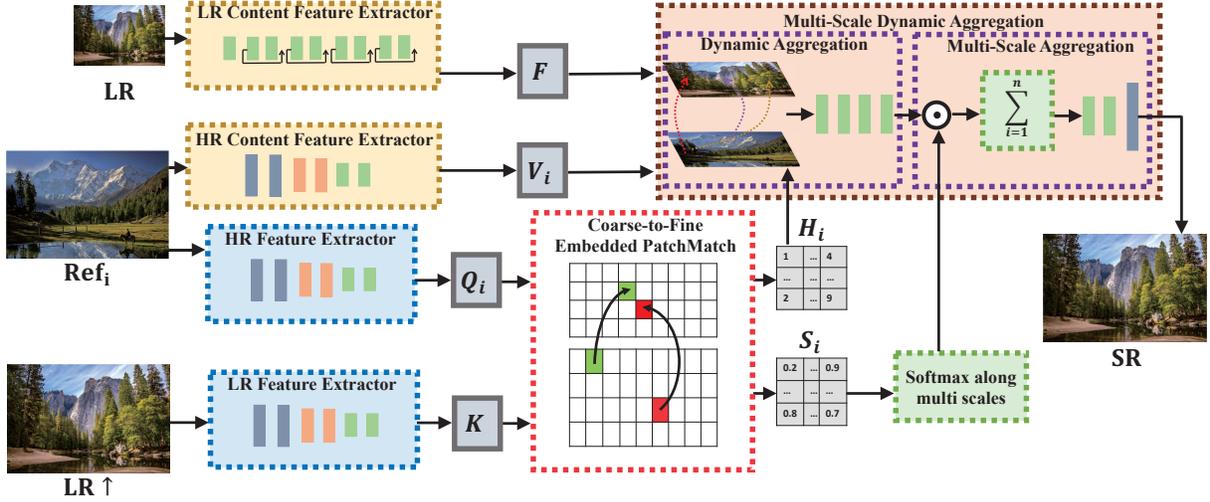} 
	\caption{The overview of Accelerated Multi-Scale Aggregation network for Reference-based Super-Resolution.  } 
	\label{fig:all_process}
\end{figure*}
The overview of our proposed AMSA is shown in Figure \ref{fig:all_process}. Given the LR and Ref image, we apply 4$\times$ bicubic-upsampling on LR to obtain  LR$\uparrow$ image and $k^{i}\times$ bicubic-downsampling on Ref image to obtain multi-scale reference images Ref$_{i}$, where $k$ ($k<1$) is a constant and $i \in [0,n)$. For correspondence matching, HR Feature Extractor and LR Feature Extractor, pretrained with contrastive learning \cite{he2020momentum} and knowledge distillation \cite{hinton2015distilling} as $C^{2}$-Matching did, extracts feature maps $\boldsymbol{Q}_{i}$ and $\boldsymbol{K}$ from Ref$_{i}$ and LR$\uparrow$ image respectively. Afterward, CFE-PatchMatch matches correspondences between $\boldsymbol{Q}_{i}$ and $\boldsymbol{K}$ efficiently and effectively, obtaining correspondence position map $\boldsymbol{H}_{i}$ and correspondence relevance map $\boldsymbol{S}_{i}$. For SR restoration, LR and HR Content Feature Extractor extract feature map $\boldsymbol{F}$ and $\boldsymbol{V}_{i}$ from LR and Ref$_{i}$ for relevant features transfer. After that, to alleviate minor scale misalignment between LR and Ref images, the Dynamic Aggregation of MSDA dynamically aggregates features from $\boldsymbol{V}_{i}$ and $\boldsymbol{F}$ according to $\boldsymbol{H}_{i}$. To fully exploit reference information across multiple scales and further enhance robustness to large scale misalignment, the Multi-Scale Aggregation of MSDA fuses all features maps according to $\boldsymbol{S}_{i}$ and uses several convolutional layers to generate SR results.

\subsection{Coarse-to-Fine Embedded PatchMatch }

\begin{figure}[h]
	\centering
	\includegraphics[height=6cm,width=8cm]{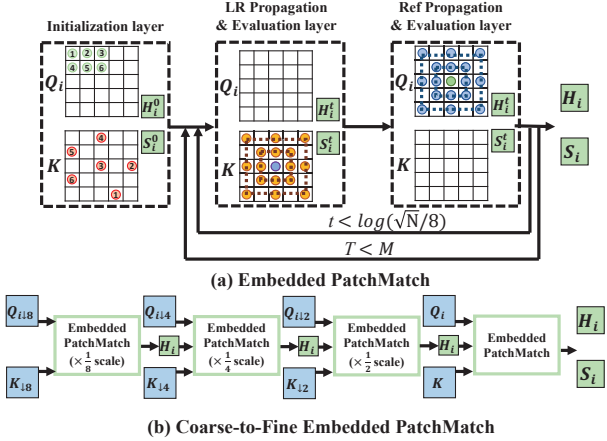} 
	\caption{ The illustration of CFE-PatchMatch. (a) Embedded PatchMatch repeats $M\log(\sqrt{N}/8)$ times ($t=$ 0 to $\log(\sqrt{N}/8)-1$, $T=$ 0 to $M-1$) to predict the correspondence position map, where $N$ is the Ref and LR$\uparrow$ image spatial size. $\boldsymbol{H}_{i}^{t}$ and $\boldsymbol{S}_{i}^{t}$ are the middle correspondence positionx and relevance relevance in $t$-th iteration. (b) Coarse-to-Fine Embedded PatchMatch applies Embedded Patchmatch on $\frac{1}{8}$, $\frac{1}{4}$, $\frac{1}{2}$ and original scales to obtain $\boldsymbol{H}_{i}$ and $\boldsymbol{S}_{i}$ rapidly.}
	\label{fig:CFE-PatchMatch}
\end{figure} 
Previous RefSR methods \cite{{TTSR},{SRNTT},{C2Matching}} match relevant patches by the enumerated global searching between LR and Ref images consuming massive computational resources. Although MASA \cite{MASA} matches correspondences in the coarser scale to reduce the computational cost, its complexity is still quadratic with respect to the input size. To tackle this issue, we design a Coarse-to-Fine Embedded PatchMatch (CFE-PatchMatch) scheme to further reduce the computational cost with a hierarchical random matching strategy.

\textbf{Embedded PatchMatch.}  Embedding PatchMatch is embedded in the network participating in end-to-end training. The illustration of Embedded PatchMatch is shown in Figure \ref{fig:CFE-PatchMatch} (a). The layers of Embedded PatchMatch are designed as follows: 

\begin{enumerate}
	\item\textbf{Initialization layer.} As shown in Figure \ref{fig:CFE-PatchMatch} (a), for initialization, the patches in $\boldsymbol{Q}_{i}$ match patches in $\boldsymbol{K}$ randomly or according to previously obtained the correspondence position map $ \boldsymbol{H}_{i}^{0} \in {\rm \mathbb{R}^{ H_{\boldsymbol{K}}\times W_{\boldsymbol{K}}}}$. Then we calculate $ \boldsymbol{S}_{i}^{0} \in {\rm \mathbb{R}^{H_{\boldsymbol{K}}\times W_{\boldsymbol{K}}}}$, which represents the relevance between LR$\uparrow$ and Ref$_{i}$ images. For each patch position $j$ in $\boldsymbol{K}$ and $j^{\prime}$ in $\boldsymbol{Q}_{i}$, we calculate the relevance $r_{i,j,j^{\prime}}$ between these two patches by normalized inner product, which is expressed as Eq \ref{equal:V1}. Consequently, given patch position $j$ and its matched patch position $\boldsymbol{H}_{i,j}^{0}$ in $\boldsymbol{Q}_{i}$, we obtain the relevance $\boldsymbol{S}_{i,j}^{0}$ as Eq \ref{equal:V2}:
	\begin{equation}
	r_{i,j,j^{'}} = \left\langle \frac{\boldsymbol{K}_{j}}{{\left\| {{\boldsymbol{K}_{j}}}
			\right\|}_2}, \frac{\boldsymbol{Q}_{i,j^{\prime}}}{{\left\| {{\boldsymbol{Q}_{i,j^{\prime}}}}
			\right\|}_2}\right\rangle,
	\label{equal:V1}
	\end{equation}
	\begin{equation}
	\boldsymbol{S}_{i,j}^{0} = r_{i,j,\boldsymbol{H}_{i,j}^{0}}.
	\label{equal:V2}
	\end{equation}
	
	\item\textbf{LR Propagation \& Evaluation layer.}  Each patch position $j$ in $\boldsymbol{K}$ propagates to its eight neighbors with dilation $2^{t}$. Then we compute the relevance between $\boldsymbol{K}_{j}$ and the matched patch of its neighbors and update with the most relevant one. The process can be expressed as:   
	\begin{equation}
	\boldsymbol{S}_{i,j}^{t+1}= \mathop{\max}_{{u\in N_{\boldsymbol{K}}(j)}} r_{i,j,\boldsymbol{H}_{i,u}^{t}},
	\end{equation}
	
	\begin{equation}
	u^{\prime} = \mathop{\arg \max}_{{u\in N_{\boldsymbol{K}}(j)}} r_{i,j,\boldsymbol{H}_{i,u}^{t}},
	\end{equation}
	\begin{equation}
	\boldsymbol{H}_{i,j}^{t+1} = \boldsymbol{H}_{i,u^{\prime}}^{t},
	\end{equation}
	where $N_{\boldsymbol{K}}(j)$ is a set including position $j$ and its eight neighbors with dilation $2^t$.
	
	\item\textbf{Ref Propagation \& Evaluation layer.} The matched patch position $\boldsymbol{H}_{i,j}$ in $\boldsymbol{Q}_{i}$ of each patch position $j$ in $\boldsymbol{K}$ propagates to eight neighbors with dilation $2^t$. Then we compute the relevance between $\boldsymbol{K}_{j}$ and the neighbors of $\boldsymbol{H}_{i,j}$, and update with the best-matched patch as step 2 does.
	
	\item\textbf{Iteration.} Repeat step 2 and step 3 $M\log(\sqrt{N}/8)$ times.
	
\end{enumerate} 

\textbf{Coarse-to-Fine Embedded PatchMatch.}  The illustration of CFE-PatchMatch is shown in Figure \ref{fig:CFE-PatchMatch} (b).  CFE-PatchMatch performs the basic Embedded PatchMatch module on multiple resolutions to predict correspondence position map H$_{i}$ while further reduce computational cost and accelerate convergence.

\textbf{Computational Complexity Analysis.} The Embedded PatchMatch iterates $M\log(\sqrt{N}/8)$ steps, where $M$ is a constant, and $N$ is the Ref and LR$\uparrow$ image spatial size.  Each patch of Embedded PatchMatch propagates to eight neighbors with dilation $2^t$ and repeats $M\log(\sqrt{N}/8)$ times, the computational complexity of which is $\mathcal{O}(MN\log(\sqrt{N}/8))$. The Coarse-to-Fine Embedded PatchMatch, as shown in Figure \ref{fig:CFE-PatchMatch} (b), applies Embedded PatchMatch on $1/8$, $1/4$, $1/2$, and $1$ scales, which consumes around $\mathcal{O}(\frac{85}{64}MN\log(\sqrt{N}/8)-\frac{27}{64}MN)$.

\subsection{Multi-Scale Dynamic Aggregation}  
To alleviate the performance drop caused by scale misalignment between the LR and Ref image and make full use of reference information across multiple scales, we propose Multi-Scale Dynamic Aggregation (MSDA), including Dynamic Aggregation and Multi-Scale Aggregation. In Dynamic Aggregation, we adopt the modified DCN form \cite{Deform} to correct minor scale misalignment yet fail to deal with large scale misalignment. Thus, we design a Multi-Scale Aggregation by aggregating multi-scale reference information to obtain robustness to large scale misalignment.

\textbf{Dynamic Aggregation.} After obtaining correspondence position map $\boldsymbol{H}_{i}$ and relevance map $\boldsymbol{S}_{i}$ between LR$\uparrow$ and Ref$_{i}$ images, MSDA dynamically aggregate reference features across multiple scales. Specifically, for each position $p$ in $\boldsymbol{K}$, we need to aggregate the features around its correspondence position $\boldsymbol{H}_{i,p}$ in $\boldsymbol{Q}_{i}$. Thus, the offset between position $p$ and $\boldsymbol{H}_{i,p}$ can be denoted as $p_{0}=\boldsymbol{H}_{i,p}-p$. The dynamic aggregation can be expressed as a modified DCN form:
\begin{equation}
\boldsymbol{Y}^{\prime}_{i}(p)=\sum_{j=1}^{9} w_{j} \cdot \boldsymbol{V}_{i}\left(p+p_{0}+p_{j}\right),
\end{equation}
\begin{equation}
\Delta \boldsymbol{P}=\operatorname{Conv}\left(\left[ \boldsymbol{F} ; \boldsymbol{Y}_{i}^{\prime}\right]\right),
\end{equation}
\begin{equation}
\boldsymbol{Y}_{i}(p)=\sum_{j=1}^{9} w_{j} \cdot \boldsymbol{V}_{i}\left(p+p_{0}+p_{j}+\Delta \boldsymbol{P}_{j}(p)\right),
\end{equation}
where $w_{j}$ denotes the convolution kernel weight, and $p_{j} \in \{(1,1),(1,0)...(-1,-1)\}$. $\boldsymbol{Y}_{i}^{\prime}$ and $\boldsymbol{Y}_{i}$ are the standard aggregated reference feature and dynamic aggregated reference feature, respectively. $\Delta \boldsymbol{P}$ is the dynamic offset map, and $\boldsymbol{F}$ is the feature map of LR. In addition, $\operatorname{Conv}$ indicates the convolution operation, and $\left[ ; \right]$ represents the concatenation operation.

\textbf{Multi-Scale Aggregation.} To enhance the robustness to large scale misalignment and fully utilize reference information $\boldsymbol{Y}_{i}$ across scales, we fuse multi-scale reference features by relevance map $\boldsymbol{S}_{i}$. The process is expressed as :
\begin{equation}
\boldsymbol{S}_{i}^{\prime}=\frac{\exp \left(\boldsymbol{S}_{i}\right)}{\sum_{j=1}^{n} \exp \left(\boldsymbol{S}_{j}\right)},
\end{equation}
\begin{equation}
\boldsymbol{Z}=\operatorname{Conv} \left(\sum_{i}^{n} \boldsymbol{Y}_{i} \cdot \boldsymbol{S}_{i}^{\prime}\right),
\end{equation}
where $\boldsymbol{Z}$ is the output SR image, and $n$ is the number of downsampled Ref images.
%

\subsection{Loss Functions}

We use three commonly used loss functions to train our model, including reconstruction loss ${\cal L}_{rec}$, perceptual loss ${\cal L}_{per}$, and adversarial loss ${\cal L}_{adv}$. For the reconstruction loss, we adopt the $\ell_{1}$-norm. For the perceptual loss, we calculate it on relu5-1 VGG \cite{vgg} features. For the adversarial loss, we employ WGAN-GP \cite{WGAN-GP}. The overall loss function of our model is ultimately designed as:

\begin{equation}
{\cal L} = \lambda_{rec} {\cal L}_{rec}+\lambda_{per} {\cal L}_{per}+\lambda_{adv} {\cal L}_{adv}.
\end{equation} 

\begin{table*}[htbp]
	\centering
	\caption{\textbf{Quantitative Comparisons}. PSNR / SSIM are used for evaluation. We group methods by SISR and RefSR. We mark the best
		results in bold. The models trained with GAN loss are marked in gray. The suffix '\textit{rec}' means only reconstruction loss is used for training.}
	\begin{tabular}{c|c|cccc}
		\hline
		& Method & CUFED5 & Sun80 & Urban100 & Manga109 \\
		\hline
		\hline
		& SRCNN & 25.33 / .745 & 28.26 / .781 & 24.41 / .738 & 27.12 / .850 \\
		& EDSR  & 25.93 / .777 & 28.52 / .792 & 25.51 / .783 & 28.93 / .891 \\
		& RCAN  & 26.06 / .769 & 29.86 / .810 & 25.42 / .768 & 29.38 / .895 \\
		SISR & \cellcolor[rgb]{ .906,  .902,  .902}SRGAN  & \cellcolor[rgb]{ .906,  .902,  .902}24.40 / .702 & \cellcolor[rgb]{ .906,  .902,  .902}26.76 / .725 & \cellcolor[rgb]{ .906,  .902,  .902}24.07 / .729 & \cellcolor[rgb]{ .906,  .902,  .902}25.12 / .802 \\
		& ENet  & 24.24 / .695 & 26.24 / .702 & 23.63 / .711 & 25.25 / .802 \\
		& \cellcolor[rgb]{ .906,  .902,  .902}ESRGAN & \cellcolor[rgb]{ .906,  .902,  .902}21.90 / .633 & \cellcolor[rgb]{ .906,  .902,  .902}24.18 / .651 & \cellcolor[rgb]{ .906,  .902,  .902}20.91 / .620 & \cellcolor[rgb]{ .906,  .902,  .902}23.53 / .797 \\
		& \cellcolor[rgb]{ .906,  .902,  .902}RankSRGAN & \cellcolor[rgb]{ .906,  .902,  .902}22.31 / .635 & \cellcolor[rgb]{ .906,  .902,  .902}25.60 / .667 & \cellcolor[rgb]{ .906,  .902,  .902}21.47 / .624 & \cellcolor[rgb]{ .906,  .902,  .902}25.04 / .803 \\
		\hline
		\hline
		& CrossNet  & 25.48 / .764 & 28.52 / .793 & 25.11 / .764 & 23.36 / .741 \\
		& \cellcolor[rgb]{ .906,  .902,  .902}SRNTT & \cellcolor[rgb]{ .906,  .902,  .902}25.61 / .764 & \cellcolor[rgb]{ .906,  .902,  .902}27.59 / .756 & \cellcolor[rgb]{ .906,  .902,  .902}25.09 / .774 & \cellcolor[rgb]{ .906,  .902,  .902}27.54 / .862 \\
		& SRNTT-\textit{rec} & 26.24 / .784 & 28.54 / .793 & 25.50 / .783 & 28.95 / .885 \\
		& \cellcolor[rgb]{ .906,  .902,  .902}TTSR & \cellcolor[rgb]{ .906,  .902,  .902}25.53 / .765 & \cellcolor[rgb]{ .906,  .902,  .902}28.59 / .774 & \cellcolor[rgb]{ .906,  .902,  .902}24.62 / .747 & \cellcolor[rgb]{ .906,  .902,  .902}28.70 / .886 \\
		& TTSR-\textit{rec}  & 27.09 / .804 & 30.02 / .814 & 25.87 / .784 & 30.09 / .907 \\
		& \cellcolor[rgb]{ .906,  .902,  .902}SSEN & \cellcolor[rgb]{ .906,  .902,  .902}25.35 / .742 & \cellcolor[rgb]{ .906,  .902,  .902}- & \cellcolor[rgb]{ .906,  .902,  .902} - & \cellcolor[rgb]{ .906,  .902,  .902}- \\
		RefSR & SSEN-\textit{rec}  & 26.78 / .791 & -     & -     & - \\
		 & \cellcolor[rgb]{ .906,  .902,  .902}CIMR & \cellcolor[rgb]{ .906,  .902,  .902}26.16 / .781 & \cellcolor[rgb]{ .906,  .902,  .902}29.67 / .806 & \cellcolor[rgb]{ .906,  .902,  .902}25.24 / .778 & \cellcolor[rgb]{ .906,  .902,  .902}- \\
		 & CIMR-\textit{rec}  & 26.35 / .789 & 30.07 / .813 & 25.77 / .792 & - \\
		& \cellcolor[rgb]{ .906,  .902,  .902}MASA & \cellcolor[rgb]{ .906,  .902,  .902}24.92 / .729 & \cellcolor[rgb]{ .906,  .902,  .902}27.12 / .708 & \cellcolor[rgb]{ .906,  .902,  .902}23.78 / .712 & \cellcolor[rgb]{ .906,  .902,  .902}- \\
		& MASA-\textit{rec}  & 27.54 / .814 & 30.15 / .815 & 26.09 / .786 &  - \\
		& \cellcolor[rgb]{ .906,  .902,  .902} $C^{2}$-Matching & \cellcolor[rgb]{ .906,  .902,  .902}27.16 / .805 & \cellcolor[rgb]{ .906,  .902,  .902}29.75 / .799 & \cellcolor[rgb]{ .906,  .902,  .902}25.52 / .764 & \cellcolor[rgb]{ .906,  .902,  .902}29.73 / .893 \\
		&  $C^{2}$-Matching-\textit{rec} &	 28.24 / .841 &	
		30.18 / .817 &	 26.03 / .785 & 30.47 / .911 \\
		\hline
		Ours  & \cellcolor[rgb]{ .906,  .902,  .902}AMSA & \cellcolor[rgb]{ .906,  .902,  .902}27.31 / .809 & \cellcolor[rgb]{ .906,  .902,  .902}29.83 / .803 & \cellcolor[rgb]{ .906,  .902,  .902}25.60 / .770 & \cellcolor[rgb]{ .906,  .902,  .902}29.79 / .896 \\
		& AMSA-\textit{rec} & \textbf{28.50} / \textbf{.849} & \textbf{30.29} / \textbf{.819} & \textbf{26.18} / \textbf{.789} & \textbf{30.57} / \textbf{.914} \\
		\hline
	\end{tabular}%
	\label{tab:quantitative}%
\end{table*}

\begin{figure*}[htb]
	\centering
	\includegraphics[height=10.2cm,width=16cm]{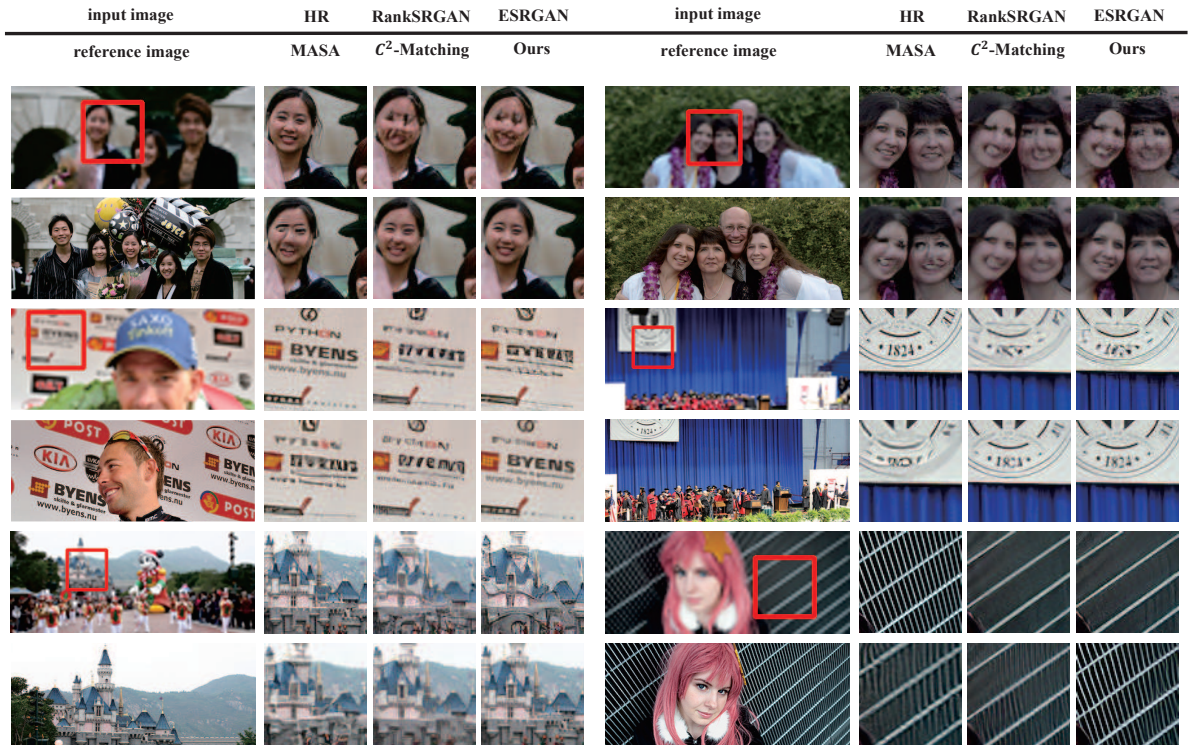} 
	\caption{\textbf{Qualitative Comparisons}. For all the shown examples, our method significantly outperforms other state-of-the-arts, particularly in the image rich in texture details.}
	\label{fig:exp}
\end{figure*}

\section{Experiments}
\subsection{Dataset}

\begin{figure}
	\vskip -5ex
	\centering
	\includegraphics[height=5.25cm,width=7cm]{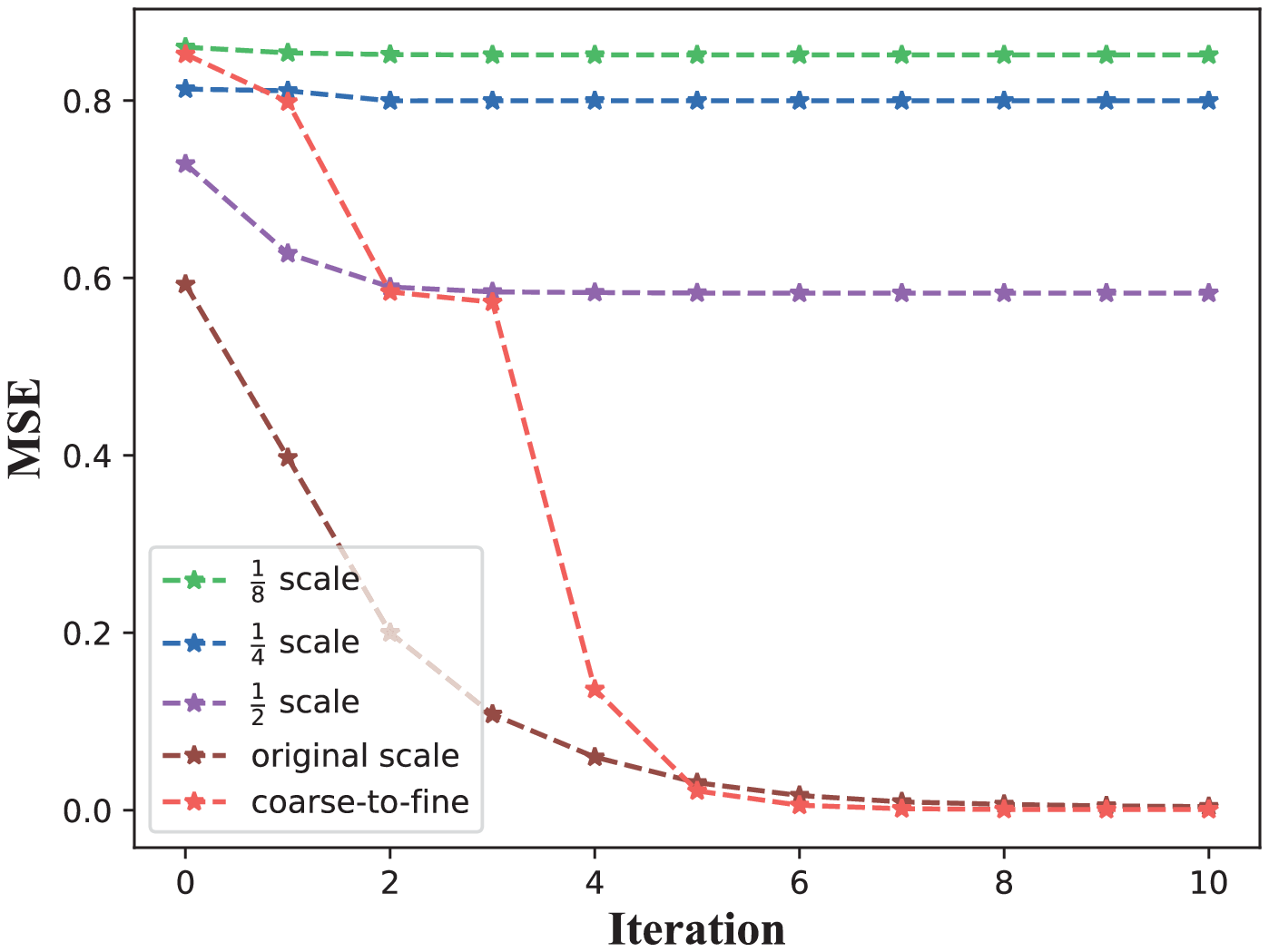} 
	\caption{Exploration on Coarse-to-Fine Embedded PatchMatch convergence in different scales. The horizontal axis represents the number of iteration times, and the vertical axis indicates the Mean Squared Error (MSE) between the CFE-PatchMatch and enumerated matching results. }
	\label{fig:patchmatch_exp}
\end{figure}

We train and test our network on the CUFED5 \cite{SRNTT} dataset. The CUFED5 dataset contains a training set with 11871 $160 \times 160$ image pairs and a testing set with 126 images accompanied by 5 reference images in different relevance levels. Additionally, we test our network on the Sun80, Urban100, and Manga109 datasets. The Sun80 \cite{sun80} dataset consists of 80 natural images, each of which includes several reference images. The Urban100 \cite{selfEX} dataset without reference images contains 100 architectural images with strong self-similarity, and the LR versions of images serve as reference images. Manga109 \cite{MANGA109} also lacks reference images, and we randomly select HR reference images from this dataset.

\subsection{Implementation Details}

For Coarse-to-Fine Embedded PatchMatch, the iteration $M$ of Embedded PatchMatch on $\frac{1}{8}$, $\frac{1}{4}$, $\frac{1}{2}$ and original scales are set to 1, 1, 2, and 6 times, separately. In addition, for the MSDA module of AMSA, we set the downsampled factor $k$ and the number of downsampled $Ref_{i}$ to 0.8 and 5, respectively. 

AMSA is trained and tested in a scale factor of 4 between the LR and HR image. We augment training data by randomly horizontally and vertically flipping followed by randomly rotating 90$^{\circ}$, 180$^{\circ}$, and 270$^{\circ}$. The model is optimized by ADAM optimizer with $\beta_{1}= 0.9$, $\beta_{2}=0.99$ and initial learning rate of 1e-4. Each mini-batch includes 9 LR patches with size $40\times40$ along with 9 Ref patches with size $160\times160$. The weights for ${\cal L}_{rec}$, ${\cal L}_{per}$, and ${\cal L}_{adv}$ are 1.0, 10-4, and 10-6, respectively. The model is implemented by PyTorch on an NVIDIA 2080Ti GPU.

\subsection{Comparison with State-of-the-Art Methods}

We compare the proposed AMSA with state-of-the-art SISR and RefSR algorithms. The compared SISR methods include SRCNN \cite{SRCNN}, EDSR \cite{EDSR}, RCAN \cite{RCAN}, SRGAN \cite{SRGAN}, ENet \cite{Enet}, ESRGAN \cite{ESRGAN}, RankSRGAN \cite{RSRGAN}, among which RCAN has achieved superior performance in terms of PSNR/SSIM. Thanks to the generative adversarial learning strategy, ESRGAN and RSRGAN gain state-of-the-art visual quality. For RefSR, the compared methods include CrossNet \cite{CrossNet}, SRNTT \cite{SRNTT}, TTSR \cite{TTSR}, SSEN \cite{SSEN}, CIMR \cite{CIMR-SR}, MASA \cite{MASA}, and $C^{2}$-Matching \cite{C2Matching}, among which $C^{2}$-Matching and MASA outperform the previous RefSR methods.

\textbf{Quantitative evaluation.}  For fair comparison with other methods mainly minimizing MAE or MSE on PSNR and SSIM, we train AMSA-\textit{rec} with only reconstruction loss function. The results of the quantitative evaluation are illustrated in Table \ref{tab:quantitative}. Our AMSA outperforms leading methods by a large margin on four datasets. Specifically, compared with $C^2$-Matching-$\textit{rec}$, MASA-$\textit{rec}$ and RCAN, the proposed AMSA-$\textit{rec}$ improves over 0.25 dB in the CUFED5 benchmark while over 0.1 dB in Sun80, Urban100, and Manga109 datasets. The results demonstrate that our method is superior to other SR methods.

\textbf{Qualitative evaluation.}  The results of the qualitative evaluation are shown in Figure \ref{fig:exp}, and our method has the best visual quality containing many realistic details close to respective HR ground-truths. Specifically, as shown in the first line, AMSA recovers the challenging face details successfully compared with other methods. Furthermore, as shown in the second and third lines, AMSA achieves significant improvements in restoring the word and architecture textures.

\section{Ablation Study}

\begin{figure}
	\centering
	\includegraphics[height=2.9cm,width=8cm]{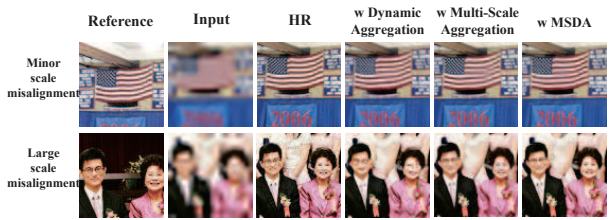} 
	\caption{Ablation study on the MSDA module.}
	\label{fig:MSDA}
\end{figure}

\begin{table*}[htbp]
	\centering
	\caption{Ablation experiments conducted on CUFED5 to
		study the effectiveness of the proposed Multi-Scale Dynamic Aggregation (MSDA) module and CFE-PatchMatch. The middle three options are correspondence matching modules, where the Enumerated-Matching is adopted in $C^{2}$-Matching, and Match $\&$ Extraction Module (MEM) is the matching acceleration module proposed in MASA.}
	\begin{tabular}{m{2cm}<{\centering} | m{2cm}<{\centering} m{2cm}<{\centering} m{2cm}<{\centering} | m{2cm}<{\centering} m{2cm}<{\centering}}
		\hline
		 MSDA & CFE-PatchMatch &Enumerated Matching & MEM   & PSNR  & GFLOPs \\
		\hline
		\hline
		\Checkmark     & \Checkmark     & \XSolidBrush     & \XSolidBrush     & 28.50 & 279.52 \\
		\hline
		\Checkmark     & \XSolidBrush     & \Checkmark     & \XSolidBrush     & 28.51 & 29297.82 \\
		\hline
		\Checkmark     & \XSolidBrush     & \XSolidBrush     & \Checkmark     & 28.43 & 1479.87 \\
		\hline
		\hline
		\XSolidBrush     & \Checkmark     & \XSolidBrush     & \XSolidBrush     & 28.21 & 83.15 \\
		\hline
		\XSolidBrush     & \XSolidBrush     & \Checkmark     & \XSolidBrush     & 28.24 & 8715.44 \\
		\hline
		\XSolidBrush     & \XSolidBrush     & \XSolidBrush     & \Checkmark     & 28.09 & 440.23 \\
		\hline
	\end{tabular}%
\label{ablation}
\end{table*}%

\textbf{Accelerated Multi-Scale Aggregation network for Reference-based Super-Resolution.} To demonstrate the effectiveness of the proposed Multi-Scale  Dynamic Aggregation module (MSDA) and Coarse-to-Fine Embedded PatchMatch (CFE-PatchMatch), we progressively add modules and compare with other matching methods. The input is assumed to be the size of 250$\times$250, and channels are 256. As shown in Table \ref{ablation}, comparing the first and fourth lines, MSDA brings around 0.3 dB improvement on the PSNR. Besides, comparing the first and second lines, our CFE-PatchMatch is about 100 times more efficient than the enumerated matching method while achieving comparable performance. Comparing the first and third lines, CFE-PatchMatch is more efficient and effective than MEM.

\textbf{The convergence of Coarse-to-Fine Embedded PatchMatch in different scales.}  To explore the convergence of CFE-PatchMatch, we apply Embedded PatchMatch on $\frac{1}{8}$, $\frac{1}{4}$, $\frac{1}{2}$, and original scales, respectively. Besides, we verify the convergence of CFE-PatchMatch, which iterates 1, 1, 2, and 6 times on $\frac{1}{8}$, $\frac{1}{4}$, $\frac{1}{2}$, and original scales separately. As shown in Figure \ref{fig:patchmatch_exp}, coarse-to-fine strategy brings fast convergence with less computational cost. Specifically, CFE-PatchMatch and Embedded PatchMatch both converge in 10 times iterations on the original scale, yet CFE-PatchMatch performs the top four iterations on the coarse scale to save around half computational cost and achieve less MSE.

\begin{figure}
	\centering
	\includegraphics[height=5.25cm,width=7cm]{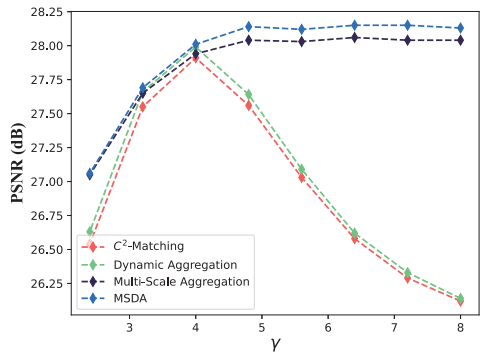} 
	\caption{The robustness for scale misalignment. The $\gamma$ indicates the scale difference between LR and reference images.}
	\label{fig:gama}
\end{figure}

\textbf{Multi-Scale Dynamic Aggregation module.} To verify the effectiveness of our MSDA to scale misalignment, we compare Dynamic Aggregation, Multi-Scale Aggregation, MSDA (Dynamic Aggregation $+$ Multi-Scale Aggregation), and $C^2$-Matching on CUFED5 (4$\times$) by downsampling Reference images with different scale factors and $\frac{1}{2}\times$ downsampling LR images. As shown in Figure \ref{fig:gama}, comparing our MSDA and  $C^2$-Matching, when the difference of scale factors ($\gamma$) between LR and reference images are larger than 4, our MSDA accurately utilizes reference details in different scales to achieve robust and superior performance.  As $\gamma$ is less than 4, reference images fail to provide  $4\times$ reference details for LR images causing a drop in performance, yet MSDA still performs better than $C^2$-Matching by using several reference details in different scales. As shown in Figure \ref{fig:MSDA} and Figure \ref{fig:gama}, comparing Dynamic Aggregation and Multi-Scale Aggregation, we can see that Dynamic Aggregation performs better on minor scale misalignment, while Multi-Scale Aggregation shows superiority on large scale misalignment. 

\begin{table}
	\begin{center}
		\caption{Ablation study on the influence of the relevance between LR and Ref images.}
		\resizebox{0.8\linewidth}{!}{
			\begin{tabular}{c|c|c}
				\hline
				Method&TTSR-\textit{rec}  & MASA-\textit{rec}\\
				\hline
				\hline
				L1&  27.09/.804 &27.35/.814 \\
				\hline
				L2&   26.74/.791&26.92/.796 \\
				\hline
				L3& 26.64/.788&26.82/.793 \\
				\hline
				L4&26.58/.787&26.74/.790 \\
				\hline
				LR&26.43/.782&26.59/.784 \\
				\hline
				\hline
				Method& $C^{2}$-Matching-\textit{rec} & AMSA-\textit{rec} (ours) \\
				\hline
				\hline
				L1&  28.24/.841 &28.58/849\\
				\hline
				L2&   27.39/.813&27.52/.816\\
				\hline
				L3& 27.17/.806&27.25/.809\\
				\hline
				L4& 26.94/.799&27.04/.803\\
				\hline
				LR& 26.52/.784&26.63/.789\\
				\hline
				
		\end{tabular}
	}
		\label{simility}
	\end{center}
\end{table}

\textbf{Reference relevance influence.} To explore the influence of the relevance between the LR and Ref image on the result, we specially conduct experiments on CUFED5, which has reference images with different relevances. In Table \ref{simility}, "L1" to "L4" are reference images in CUFED5, where "L1" is the most relevant one while "L4" is the least relevant one. "LR" represents taking the LR image as Ref image. As shown in Table \ref{simility}, our AMSA achieves the best performance among RefSR methods with the same relevance level Ref image.

\section{Conclusion}
In this paper, we propose a novel Accelerated Multi-Scale Aggregation network for Reference-based Super-Resolution to enable effective and efficient reference information matching and aggregating. To reduce the massive computational cost of matching correspondences between the LR and reference image, we propose a Coarse-to-Fine Embedded PatchMatch involving end-to-end training. Furthermore, to fully exploit reference details across multiple scales and enhance robustness to scale misalignment, we develop a Multi-Scale Dynamic Aggregation module, which consists of Dynamic Aggregation and Multi-Scale Aggregation. Specifically, we apply Dynamic Aggregation to correct minor scale misalignment and employ Multi-Scale Aggregation to enhance robustness to large scale misalignment. Our method achieves state-of-the-art results both quantitatively and qualitatively on different datasets.

\section{Acknowledgments}
This work was partly supported by the Natural Science Foundation of China (No.62171251), the Natural Science Foundation of Guangdong Province (No.2020A1515010711), the Special Foundation for the Development of Strategic Emerging Industries of Shenzhen (No.JCYJ20200109143010272) and Oversea Cooperation Foundation of Tsinghua Shenzhen International Graduate School.

\appendix

\bibliography{aaai22}

\end{document}